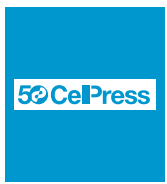

Contents lists available at ScienceDirect

# Heliyon

journal homepage: www.cell.com/heliyon

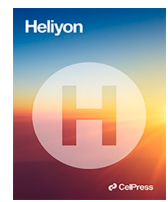

Research article

# Differentiating viral and bacterial infections: A machine learning model based on routine blood test values

Gregor Gunčar [a,c,1], Matjaž Kukar [a,b,1], Tim Smole [a], Sašo Moškon [a], Tomaž Vovko [d], Simon Podnar [e], Peter Černelč [a], Miran Brvar [f], Mateja Notar [a], Manca Köster [a], Marjeta Tušek Jelenc [a], Žiga Osterc [a], Marko Notar [a,*]

[a] *Smart Blood Analytics Swiss SA, CH-8008, Zürich, Switzerland*
[b] *Faculty of Computer and Information Science, University of Ljubljana, Slovenia*
[c] *Faculty of Chemistry and Chemical Technology, University of Ljubljana, Slovenia*
[d] *Department of Infectious Diseases, University Medical Centre Ljubljana, Slovenia*
[e] *Division of Neurology, University Medical Centre Ljubljana, Slovenia*
[f] *Centre for Clinical Toxicology and Pharmacology, University Medical Centre Ljubljana, Slovenia*

## ABSTRACT

The growing threat of antibiotic resistance necessitates accurate differentiation between bacterial and viral infections for proper antibiotic administration. In this study, a Virus vs. Bacteria machine learning model was developed to distinguish between these infection types using 16 routine blood test results, C-reactive protein concentration (CRP), biological sex, and age. With a dataset of 44,120 cases from a single medical center, the model achieved an accuracy of 82.2 %, a sensitivity of 79.7 %, a specificity of 84.5 %, a Brier score of 0.129, and an area under the ROC curve (AUC) of 0.905, outperforming a CRP-based decision rule. Notably, the machine learning model enhanced accuracy within the CRP range of 10–40 mg/L, a range where CRP alone is less informative. These results highlight the advantage of integrating multiple blood parameters in diagnostics. The "Virus vs. Bacteria" model paves the way for advanced diagnostic tools, leveraging machine learning to optimize infection management.

## 1. Introduction

Accurate and timely differentiation between bacterial and viral infections is essential to ensure appropriate antibiotic prescribing practices and mitigate the spread of antibiotic resistance [1]. The rise of antibiotic resistance poses a major threat to global public health, as it undermines the efficacy of life-saving antibiotics and increases the risk of complications and mortality associated with common infections [2]. A key driver of antibiotic resistance is the inappropriate use of antibiotics, particularly in situations where they are not clinically indicated, such as viral infections [3].

Healthcare providers commonly employ blood tests to obtain insights into a patient's health status. Complete blood count (CBC) and C-reactive protein (CRP) are among the most frequently measured blood test parameters in clinical practice, as they can provide valuable information about a patient's immune response and inflammation levels. The most commonly studied biomarkers for distinguishing between bacterial and viral infections include CRP, procalcitonin (PCT), and various cytokines [4–6]. Among these, PCT has shown the most promise due to its higher specificity and sensitivity in differentiating both bacterial infections from viral infections and bacterial infections from other noninfective causes of systemic inflammation [7]. PCT levels tend to be markedly elevated in bacterial infections, whereas they remain low in viral infections, providing a useful clinical tool for guiding antibiotic therapy [8]. CRP, an acute-phase protein produced by the liver in response to inflammation, infection, or tissue injury [9], has also been widely used as a diagnostic biomarker to differentiate between bacterial and viral infections. CRP levels are elevated in bacterial infections compared to viral infections [10] in response to pro-inflammatory cytokines, such as interleukin-6 (IL-6), which are predominantly secreted during bacterial infections [11]. However, CRP is also a non-specific marker of inflammation and can be elevated in various non-infectious conditions, such as autoimmune diseases and cancer [12].

---

\* Corresponding author.
*E-mail address:* marko@sba-swiss.com (M. Notar).
[1] These authors contributed equally.

Unfortunately, these biomarkers are often limited by their lack of specificity, as the results can be influenced by a variety of factors, like the progression of the infection and those unrelated to the underlying infection, such as age, comorbidities, and medication use. Additionally, there is considerable overlap in the values observed in bacterial and viral infections, making it difficult to establish definitive cutoffs that can reliably distinguish between the two types of infections [9].

Recent advancements in machine learning (ML) have demonstrated the potential to revolutionize the field of diagnostic medicine [13], with applications ranging from the early detection of skin cancer [14] to the identification of pneumonia in chest X-ray images

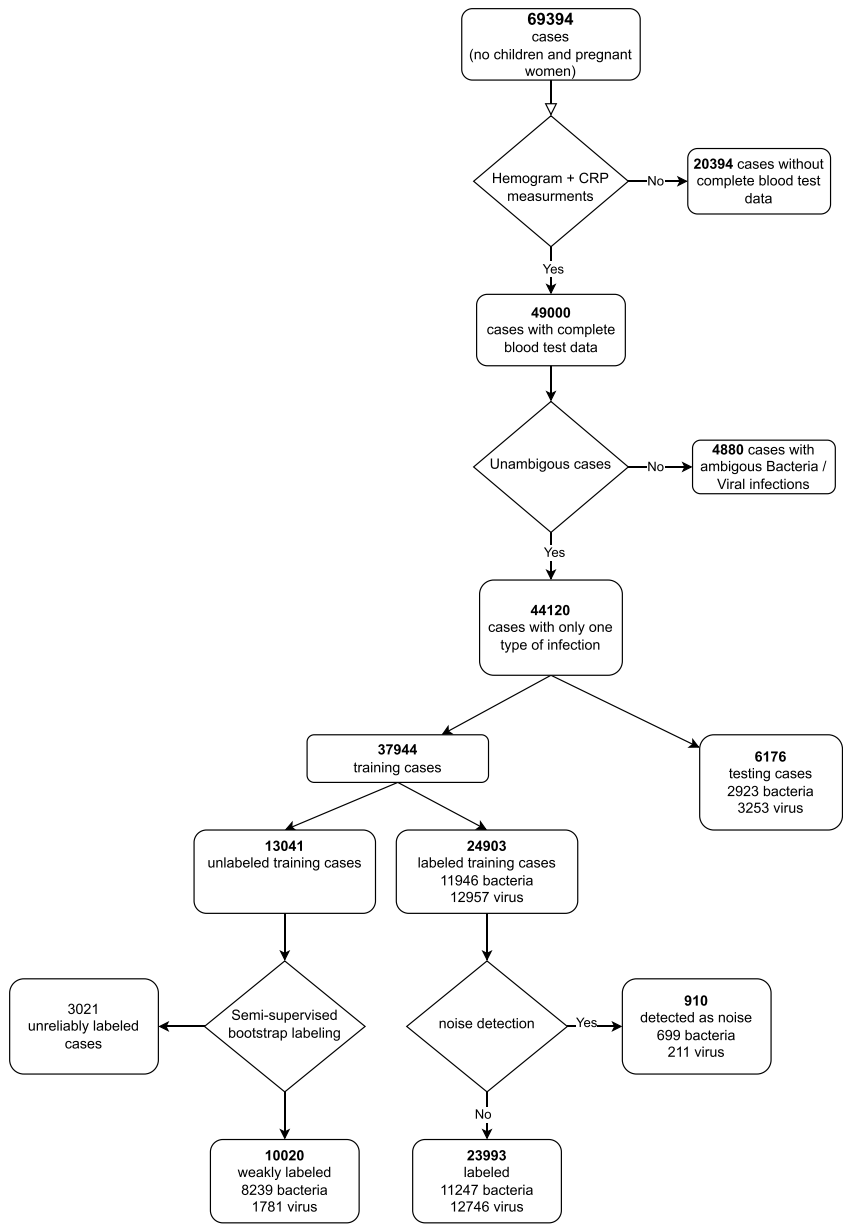

**Fig. 1.** Flow diagram of cases included in the study. Unlabeled cases were used only for training the diagnostic model. All reported results were obtained using the labeled cases only.

[15]. Machine learning models can leverage large amounts of data to uncover complex patterns and relationships that are not apparent to clinicians, enabling more accurate and efficient diagnoses [16,17]. The application of machine learning to the analysis of blood test results has the potential to improve the differentiation between bacterial and viral infections and, consequently, optimize antibiotic prescribing practices. According to several authors [18–20] who analyzed hundreds of scientific papers, the most popular supervised machine learning algorithms for learning from structured medical data are decision trees and ensembles (gradient boosting, random forests), support vector machines, nearest neighbors, and shallow artificial neural networks. For other data modalities, such as images and unstructured text, deep learning methods, such as deep neural networks, are frequently used [20].

In this study, we present a machine-learning model that differentiates between bacterial and viral infections based solely on the most frequently measured blood test values and CRP. Our model aims to provide physicians with a reliable and objective diagnostic tool that can help guide decisions for prescribing antibiotics and reduce the overuse of antibiotics when not necessary. The proposed model builds upon our previous work in the field of machine learning-based diagnostics, such as hematological disorders [21], COVID-19 [22] and brain tumors [23]. We performed a comprehensive comparison of various ML algorithms (see Model Comparison and Performance), and selected XGBoost because it achieved significantly higher classification accuracy than any other algorithm. XGBoost [24] is a leading framework for extreme gradient boosting, well respected for its scalability and performance, and widely used in medical applications.

A major focus of our model is the analysis of CRP levels, a widely used biomarker for inflammation and infection. Despite its clinical utility, the interpretation of CRP levels remains challenging due to the lack of universally accepted cutoff values for distinguishing between bacterial and viral infections [12].

Our model aims to overcome the limitations of CRP levels by adding other routinely measured blood test values, such as white blood cell counts, to improve the accuracy of differentiating between viral and bacterial infections.

## 2. Materials and methods

### 2.1. Cohort characteristics

A pool of 69,394 medical cases with various viral and bacterial infections was obtained from the population of adult, non-pregnant patients admitted to the University Medical Centre Ljubljana (UMCL), Slovenia, between January 2005 and May 2020. All cases were de-identified prior to storage and analysis in accordance with the European General Data Protection Regulation (GDPR).

We acquired data on the patient's age, sex, routine blood test results, and final diagnoses. The cases were marked by medical doctors as VIRUS, BACTERIA, or UNLABELED, where unlabeled cases were those where the cause of infection (viral or bacterial) couldn't be determined based on the ICD code.

We further excluded cases without complete blood count, or CRP measurements, resulting in 49,000 cases. In the next step, we excluded ambiguous cases where a label of another type occurred among comorbidities (i.e. cases that were labeled with viral and bacterial infection at the same time). The final number of valid cases was 44,120 (Fig. 1).

Cases were split into training and testing (evaluation) sets in a random manner, using group stratification. To account for the inter-patient dependence, which refers to the relationship between medical cases of a single patient and the class imbalance, the samples were grouped based on patient ID and stratified to ensure that all medical cases for a single patient were present in only one set and that the positive and negative class examples were balanced within each set. The training set included 37,944, of which 11,946 were bacterial, 12,957 viral infections, and 13,041 unlabeled cases. The testing (evaluation) set included 6176 cases, of which 2923 were bacterial and 3253 viral infections. Unlabeled cases were then labeled using semi-supervised bootstrap labeling and subsequently used only for training and not for model evaluation.

All methods were performed in accordance with the relevant guidelines and regulations. The National Ethics Committee of Slovenia approved the study (No. 0120–718/2015/7, No. 0120–170/2020/6, No. 0120-058/2016-2 KME 7/01/16, No. 0120–341/2016-2 KME 33/07/16, No. 0120–718/2015-2 KME 103/11/15 and No. 0120–718/2015/5); patients' written informed consent was not needed according to the Slovenian Patients' Rights act, article 44/6. The study was performed in accordance with the STARD recommendations [25].

### 2.2. Determining case diagnoses

Every medical case in our dataset (later used for building and testing the model) consisted of a final diagnosis– bacterial or viral, and 17 blood test results as the most important part of the data, accompanied by the biological sex and age of every patient. All blood test results were obtained upon admission before any treatment was introduced. All final bacterial or viral diagnoses were clinically determined by regular physician protocols using best practices for diagnosing bacterial and viral infections. This means that the Virus vs. Bacteria model has learned from physicians and, therefore, from sound medical knowledge.

### 2.3. Semi-supervised bootstrap labeling and noise detection

In many machine learning applications, including computer-aided medical diagnosis [26,27], there is often an extensive pool of unlabeled (undiagnosed) data available. In our scenario, we treated the cases where the infection could be either of bacterial or viral origin as unlabeled. We performed a semi-supervised bootstrap labeling step on the training set before training the final model (Algorithm 1 in supplemental materials). In this step, a model was built on the labeled data only – cases clearly marked as either viral or bacterial infections. This model was then used to label the unlabeled cases [27]. Only unlabeled cases that were classified with a probability higher than 70 % were assigned the predicted label and included in the final training set (Fig. 2).

Labeled cases from the training set that the model was unable to correctly predict were treated as noise. These cases (910 in total) were excluded from all training steps but were still included in the validation steps during 10-fold cross-validation. The final training set included 12,746 viral cases, 11,247 bacterial infection cases, and 10,020 cases labeled by the semi-supervised bootstrap labeling model (Table 5).

*2.4. SBAS machine learning algorithm*

The Virus vs. Bacteria model, as presented in this paper, was built using the Smart Blood Analytics Swiss (SBAS) algorithm– a CRISP-DM compliant machine learning pipeline consisting of five processing stages (Algorithm 2 in supplemental materials) corresponding to phases 2–6 of the CRISP-DM process standard [28]. The stages are as follows:

- Data acquisition: acquiring raw data from the database
- Data filtering: constructing the training dataset consisting of blood test results obtained before treatment
- Data pre-processing: the canonization of blood parameters (matching them with our reference blood parameter database, conversion to SI units, data quality control)
- Data modeling: building the diagnostic model using a ML algorithm (base learner)
- Evaluation: evaluating the model with stratified ten-fold cross-validation and/or independent testing data
- Deployment of the successfully evaluated model in the cloud (accessible either through hospital information systems or the SBAS website - https://app.smartbloodanalytics.com).

As the base ML method, any algorithm that produces probabilistic outputs can be used. We performed a comprehensive comparison of various ML algorithms (see the Base machine learning methods section) and decided on using XGBoost in production (see the Quantification and statistical evaluation section and Table S4).

*2.5. Base machine learning methods*

Each of the following machine learning methods was used and evaluated as a base learner:

- XGBoost – XGB [29] is based on gradient boosting and utilizes a combination of decision trees and regularization techniques to minimize a differentiable loss function, enabling it to effectively handle complex and high-dimensional data with typically excellent predictive performance.

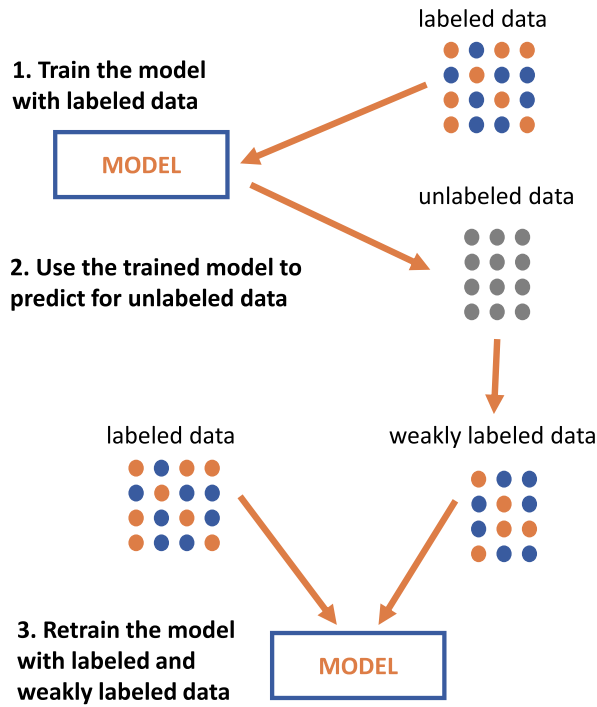

**Fig. 2.** Semi-supervised bootstrap labeling of unlabeled cases.

- Random Forest [30] is an ensemble learning algorithm that constructs a multitude of decision trees at training time and outputs the class that is the mode of the classes or mean prediction of the individual trees, achieving high accuracy and robustness to overfitting through bootstrap aggregation and feature randomness.
- K-nearest neighbors–KNN [31] classifies an example based on the majority class of its k-nearest neighbors in the feature space, using generic or custom distance metrics and efficient data structures for retrieval.
- Support Vector Machines–SVM [32] constructs a hyperplane or set of hyperplanes in a high-dimensional feature space to maximize the margin between different classes, using kernel functions to implicitly map the data into higher dimensions and achieve non-linear separation.
- Decision Trees–DT [33] recursively partitions the feature space into subsets based on the most informative feature, using a heuristic criterion to maximize the partition purity. Their structure allows for easy interpretation and visualization.
- Logistic Regression – LR [34] models the probability of a binary or multi-class outcome as a function of the input features, using a logistic function and maximum likelihood estimation to optimize the coefficients and enable efficient probabilistic classification.
- TabNet deep neural network–NN [35] is an interpretable deep learning architecture specialized for tabular data that employs sparse attention and feature masking to learn both local and global representations, achieving state-of-the-art performance on various machine learning tasks.

For KNN, SVM and LR, standard scaling has been applied by removing the mean and scaling features to unit variance. Imputation was not needed for any method since no missing values were present in the dataset.

For all methods, model parameters were optimized with the Ray Tune library using a random search [36]. Optimized models were evaluated using classification accuracy [37], sensitivity [37], specificity [37], Brier score [38], as well as AUC [37,39] and statistically compared using parametric (paired *t*-test) and non-parametric (Wilcoxon signed rank) statistical tests with Bonferroni correction [40–42].

### 2.6. Model explanation with Shapley values

Each model output can be explained by depicting Shapley values, a model-agnostic and theoretically grounded explanation framework that assigns fair and unique values to each blood parameter based on its marginal contribution to the expected output, providing a comprehensive and consistent interpretation across the models [43–45].

### 2.7. Data visualization with UMAP

The UMAP (Uniform Manifold Approximation and Projection) dimensionality reduction technique [46] was used to visualize the dataset. This method seeks to preserve the global structure of the data while reducing the dimensionality to a lower number of features that can be easily plotted on a 2D scatterplot. By projecting the high-dimensional data into a lower-dimensional space, UMAP allows for the visual inspection of the data structure and relationships between samples.

### 2.8. Model performance evaluation

The performance of a binary classifier was evaluated using grouped stratified 10-fold cross-validation. To account for the inter-patient dependence, which refers to the relationship between medical cases of a single patient, and the class imbalance, the samples were grouped based on patient ID and stratified to ensure that all medical cases for a single patient were present in only one fold at a time and that the positive and negative class examples were balanced within each fold. The classifier was trained and tested 10 times, rotating through each fold, and multiple classification metrics were calculated for each iteration, such as accuracy, precision, recall, and Brier score. The average performance was then calculated from the 10 iterations to provide an estimate of the classifier's overall ability to generalize to unseen data.

### 2.9. Baseline approach

As a baseline performance criterion, we compare our results with a simple decision rule, based on the CRP value alone. In Ref. [47] authors are skeptical of prescribing antibiotics when CRP<20 mg/L (viral infection), while in Refs. [48,49] authors suggest that for CRP>40 mg/L (or 50 mg/L), antibiotics should be prescribed (bacterial infection). As sources differ with respect to the hard threshold, above which an infection can be said to be bacterial in nature, we approached the problem optimistically by finding the optimal value for our training data.

$$CRP_{opt} = \underset{c}{\mathrm{argmax}} \{Accuracy_c(data)\}$$

Here, $Accuracy_c(data)$ stands for diagnostic accuracy of the CRP decision rule with $CRP_{opt}=c$. In our case, $CRP_{opt} = 24$ mg/L, which is not much above 20 mg/L [47]. While this value of $CRP_{opt}$ may not be the best in general, it represents the best possible (and thus optimistic) threshold for our data. The *optimal* CRP decision rule for our dataset is, therefore, as follows:

**if** $CRP < CRP_{opt}$ **then** a diagnosis is Virus **else** diagnosis is Bacteria.

Characteristics (ROC curves) of the optimal CRP decision rule are shown in Supplemental Fig. S2.

*2.10. Parameter contributions*

From Tables 2–4 we can see that diagnostic models are significantly superior to the simple CRP decision rule. As the models account for other blood parameters as well, they must consequently contribute to the final decision. In Table 1, we show descriptive statistics of blood parameters and statistical comparisons of the 'Bacteria' and 'Virus' populations using both the two-sample Anderson-Darling population test [50] and the non-parametric Mann-Whitney U (a.k.a. Wilcoxon Rank-Sum) test [51,52]. As both populations are quite large (12,957 and 11,946 cases, respectively), even small differences are statistically significant, although the differences may not always be clinically relevant. In such cases, it is best to assess the magnitude of population differences for each parameter using *Cohen's d* effect size [53]. For qualitative assessment of effect size, we use the established range of values: none (−) below 0.2, small (S) between 0.2 and 0.5, medium (M) between 0.5 and 0.8, and large (L) above 0.8. While in all parameters, the differences between populations are statistically very significant ($p < 0.00.1$), in 6 out of 18 parameters (33 %), the effect size is negligible or small.

## 3. Results

*3.1. Dataset description*

A dataset consisting of 44,120 cases was used for this study, including 37,944 cases in the training set (11,946 bacterial, 12,957 viral, and 13,041 unlabeled cases) and 6176 cases in the testing set (2923 bacterial and 3253 viral infections) (Fig. 1). They were obtained from a pool of 69,394 cases with various viral and bacterial infections admitted to the University Medical Centre Ljubljana (UMCL), Slovenia, between January 2005 and May 2020 (see details in the Methods section).

*3.2. Blood parameters (features) analysis*

For the set of blood parameters, we chose the most frequently measured and complete blood parameters in our dataset together with CRP as the most commonly used biomarker for the infection. The selected parameters are: leukocyte count, lymphocyte count, monocyte count, neutrophils count, lymphocyte %, monocyte %, neutrophils %, erythrocyte count, hemoglobin, hematocrit, mean corpuscular hemoglobin (MCH), mean corpuscular hemoglobin concentration (MCHC), mean corpuscular volume (MCV), erythrocyte distribution width (RDW), thrombocytes count, mean platelet volume (MPV), CRP and patient's age and sex. We also used the neutrophil-to-lymphocyte ratio (NLR) [54–58].

The differences in the distributions of the parameters for viral and bacterial infections were assessed using violin plots (Fig. 3) and descriptive statistics (Table 1). They both show statistically significant differences in the distribution of all blood parameters regarding viral and bacterial infections. In Table 1, we show descriptive statistics of blood parameters and statistical comparisons of the 'Bacteria' and 'Virus' populations using both the two-sample Anderson-Darling population test [50] and the non-parametric Mann-Whitney U (a. k.a. Wilcoxon Rank-Sum) test [51,52]. As both populations are quite large (12,957 cases for viral and 11,946 cases for bacterial infection), even small differences are statistically significant ($p < 0.01$), although the differences may not always be clinically relevant. We assessed the relevance of population differences for each parameter with *Cohen's d* effect size [53]. For qualitative assessment, we

**Table 1**
Descriptive statistics of blood parameters in training data set and statistical comparisons of 'Bacteria' and 'Virus' populations. The populations are statistically significantly different in all parameters using both the two-sample Anderson-Darling population test [50] and the non-parametric Mann-Whitney U (a.k.a. Wilcoxon Rank-Sum) test [51,52]. The magnitude of population differences for each parameter was assessed using Cohen's d effect size [53] on the none (−)/small (S)/medium (M)/large (L) scale.

| Parameter | Median | | | IQR | | | Anderson- Darling | Mann-Whitney U | Cohen's effect size |
|---|---|---|---|---|---|---|---|---|---|
| | All | Bacteria | Virus | All | Bacteria | Virus | (p-value) | (p-value) | (−/S/M/L) |
| **WBC [1E9/L]** | 8.1 | 10.2 | 6.4 | 5.8 | 7.1 | 3.2 | <0.01 | <0.01 | L |
| **Neutrophils count [1E9/L]** | 5.33 | 7.48 | 3.64 | 5.30 | 6.28 | 2.41 | <0.01 | <0.01 | L |
| **Lymphocyte count [1E9/L]** | 1.38 | 1.13 | 1.71 | 1.07 | 0.93 | 1.02 | <0.01 | <0.01 | M |
| **Monocyte count [1E9/L]** | 0.56 | 0.60 | 0.52 | 0.37 | 0.44 | 0.29 | <0.01 | <0.01 | S |
| **Neutrophils % [1]** | 0.693 | 0.769 | 0.592 | 0.221 | 0.163 | 0.166 | <0.01 | <0.01 | L |
| **Lymphocyte % [1]** | 0.182 | 0.113 | 0.286 | 0.197 | 0.125 | 0.145 | <0.01 | <0.01 | L |
| **Monocyte % [1]** | 0.076 | 0.067 | 0.086 | 0.043 | 0.046 | 0.037 | <0.01 | <0.01 | M |
| **RBC [1E12/L]** | 4.37 | 4.16 | 4.60 | 0.86 | 0.86 | 0.68 | <0.01 | <0.01 | M |
| **Hb [g/L]** | 132 | 124 | 140 | 27 | 26 | 21 | <0.01 | <0.01 | L |
| **Hct [1]** | 0.393 | 0.372 | 0.418 | 0.078 | 0.075 | 0.059 | <0.01 | <0.01 | L |
| **MCV [fL]** | 90.2 | 89.6 | 91.0 | 7.3 | 7.3 | 6.9 | <0.01 | <0.01 | S |
| **MCH [pg/cell]** | 30.2 | 29.9 | 30.4 | 2.5 | 2.6 | 2.4 | <0.01 | <0.01 | S |
| **MCHC [g/L]** | 333 | 333 | 333 | 14 | 16 | 13 | <0.01 | <0.01 | − |
| **RDW [1]** | 0.138 | 0.143 | 0.133 | 0.022 | 0.023 | 0.014 | <0.01 | <0.01 | M |
| **Platelet count [1E9/L]** | 210 | 227 | 193 | 108 | 128 | 84 | <0.01 | <0.01 | S |
| **MPV [fL]** | 8.5 | 8.5 | 8.5 | 1.5 | 1.5 | 1.4 | <0.01 | <0.01 | − |
| **CRP [mg/L]** | 23 | 90 | 3 | 108 | 147 | 6 | <0.01 | <0.01 | L |
| **NLR [1]** | 3.80 | 6.77 | 2.07 | 6.52 | 9.32 | 1.76 | <0.01 | <0.01 | M |
| **Age [years]** | 62 | 74 | 44 | 35 | 22 | 22 | <0.01 | <0.01 | L |

**Table 2**

10-fold cross-validation results (mean ± standard deviation). The XGBoost model (XGB) has both a significantly higher classification accuracy and a significantly lower Brier score than any other compared models. The difference between the CRP decision rule and XGB is 6.8 % ± 1.2 %.

|     | Accuracy | Sensitivity | Specificity | Brier score | AUC |
| --- | --- | --- | --- | --- | --- |
| **XGB** | 0.835 ± 0.01 | 0.796 ± 0.015 | 0.870 ± 0.013 | 0.123 ± 0.007 | 0.910 ± 0.007 |
| **KNN** | 0.820 ± 0.01 | 0.729 ± 0.017 | 0.896 ± 0.010 | 0.131 ± 0.006 | 0.896 ± 0.008 |
| **RF** | 0.830 ± 0.009 | 0.790 ± 0.013 | 0.867 ± 0.010 | 0.243 ± 0.000 | 0.904 ± 0.007 |
| **SVM** | 0.812 ± 0.007 | 0.743 ± 0.012 | 0.876 ± 0.011 | 0.137 ± 0.006 | 0.885 ± 0.007 |
| **DT** | 0.806 ± 0.01 | 0.768 ± 0.016 | 0.841 ± 0.017 | 0.145 ± 0.006 | 0.876 ± 0.007 |
| **LR** | 0.812 ± 0.008 | 0.744 ± 0.014 | 0.875 ± 0.011 | 0.137 ± 0.006 | 0.886 ± 0.007 |
| **NN** | 0.821 ± 0.01 | 0.779 ± 0.019 | 0.861 ± 0.021 | 0.130 ± 0.006 | 0.900 ± 0.007 |
| **CRP** | 0.767 ± 0.006 | 0.653 ± 0.008 | 0.872 ± 0.007 | 0.177 ± 0.003 | 0.813 ± 0.008 |

**Table 3**

10-fold stratified cross-validation results (mean ± standard deviation) on the region with CRP between 10 and 40 mg/L. The XGBoost model (XGB) has both a significantly higher classification accuracy (with the exception of RF) and a significantly lower Brier score than any other compared models. Note especially the increased difference between the CRP decision rule and XGB (20.9 % ± 3.2 %).

|     | Accuracy | Sensitivity | Specificity | Brier score | AUC |
| --- | --- | --- | --- | --- | --- |
| **XGB** | 0.760 ± 0.025 | 0.755 ± 0.045 | 0.764 ± 0.016 | 0.171 ± 0.016 | 0.834 ± 0.024 |
| **KNN** | 0.732 ± 0.029 | 0.683 ± 0.046 | 0.778 ± 0.027 | 0.187 ± 0.018 | 0.809 ± 0.027 |
| **RF** | 0.751 ± 0.034 | 0.751 ± 0.046 | 0.751 ± 0.027 | 0.245 ± 0.000 | 0.819 ± 0.027 |
| **SVM** | 0.720 ± 0.022 | 0.697 ± 0.035 | 0.741 ± 0.022 | 0.193 ± 0.013 | 0.790 ± 0.022 |
| **DT** | 0.698 ± 0.016 | 0.672 ± 0.033 | 0.721 ± 0.025 | 0.218 ± 0.012 | 0.760 ± 0.018 |
| **LR** | 0.722 ± 0.024 | 0.700 ± 0.037 | 0.741 ± 0.020 | 0.193 ± 0.013 | 0.792 ± 0.022 |
| **NN** | 0.736 ± 0.021 | 0.720 ± 0.049 | 0.750 ± 0.050 | 0.184 ± 0.015 | 0.811 ± 0.023 |
| **CRP** | 0.551 ± 0.020 | 0.463 ± 0.036 | 0.632 ± 0.025 | 0.298 ± 0.011 | 0.563 ± 0.022 |

**Table 4**

Results on an independent evaluation dataset (6176 test cases). Differences in accuracy and Brier score between the XGBoost model (XGB) and the CRP decision rule are statistically significant, albeit slightly smaller compared to 10-fold cross validation results (5.1 % ± 1.4 % vs 6.8 % ± 1.2 % for all cases and 17.9 % ± 4.1 % vs 20.9 % ± 3.2 % for the CRP range between 10 and 40 mg/L).

| (all cases) | Accuracy | Sensitivity | Specificity | Brier score | AUC |
| --- | --- | --- | --- | --- | --- |
| **XGB** | 0.822 ± 0.010 | 0.797 | 0.845 | 0.129 | 0.905 |
| **CRP** | 0.771 ± 0.010 | 0.662 | 0.869 | 0.175 | 0.818 |
| (cases with CRP 10–40 mg/L) | **Accuracy** | **Sensitivity** | **Specificity** | **Brier score** | **AUC** |
| **XGB** | 0.732 ± 0.029 | 0.678 | 0.790 | 0.180 | 0.822 |
| **CRP** | 0.553 ± 0.033 | 0.470 | 0.632 | 0.298 | 0.577 |

used the established range of values: none (−) below 0.2, small (S) between 0.2 and 0.5, medium (M) between 0.5 and 0.8, and large (L) above 0.8. The effect size is large in WBC, Hb, Hct, Neutrophils count, Neutrophils %, Lymphocyte %, CRP, and age.

Differences in both types of infections can also be visualized with the UMAP (Uniform Manifold Approximation and Projection) dimensionality reduction technique (Fig. 4).

### 3.3. Labeling

Every medical case in the dataset includes a confirmed ICD-encoded diagnosis relating to its origin: bacterial or viral infection. The diagnoses were obtained by regular physician protocols using best practices for diagnosing bacterial and viral infections. For 13,041 cases, it was - based on the ICD code only - not possible to decide whether the infection was of bacterial or viral origin. Such cases were used in a semi-supervised manner (see details in the Methods section) for training and weren't used in any evaluation.

### 3.4. Model comparison and performance

For the training of the diagnostic model, seven well-known supervised machine learning approaches were considered, described in the Base machine learning methods section. For each model, an extensive hyperparameter optimization was performed, using a random search, with the objective of maximizing the accuracy. In Table 2, we report the mean results and their standard deviations obtained from 10-fold cross-validation for the best models obtained by hyperparameter optimization. For sensitivity, specificity, ROC-curve, and related visualizations, the 'Bacteria' diagnosis is considered as positive to comply with the established presentation. For consistency, the same data partitioning into folds was used for all the experiments.

We tested the statistical significance of results using the one-sided non-parametric Wilcoxon signed rank test [40,59], as well as the parametric paired *t*-test. In both cases, to account for multiple testing (seven methods and CRP decision criterion), p-values were

**Table 5**

Dataset Summary. We are reporting median ± interquartile deviation (IQR/2) or counts (for biological sex only).

| Parameter | TRAINING SET (incl. unlabeled) | | | TESTING SET | | |
| --- | --- | --- | --- | --- | --- | --- |
| | ALL | BACTERIA | VIRUS | ALL | BACTERIA | VIRUS |
| **No. cases** | 34013 | 19486 | 14527 | 6176 | 2923 | 3253 |
| **Biological sex (male\|female)** | 18925\|15088 | 9741\|9745 | 9184\|5343 | 3627\|2549 | 1483\|1440 | 2144\|1109 |
| **Age [years]** | 62 ± 17.5 | 74 ± 11 | 44 ± 11 | 57 ± 16 | 70 ± 12.5 | 46 ± 11 |
| **WBC [1E9/L]** | 8.1 ± 2.9 | 10.2 ± 3.55 | 6.4 ± 1.6 | 7.3 ± 2.35 | 9.2 ± 3.4 | 6.3 ± 1.65 |
| **Neutrophils count [1E9/L]** | 5.33 ± 2.65 | 7.48 ± 3.14 | 3.64 ± 1.21 | 4.60 ± 2.06 | 6.40 ± 3.12 | 3.65 ± 1.22 |
| **Lymphocyte count [1E9/L]** | 1.38 ± 0.53 | 1.13 ± 0.47 | 1.71 ± 0.51 | 1.46 ± 0.54 | 1.21 ± 0.51 | 1.66 ± 0.52 |
| **Monocyte count [1E9/L]** | 0.56 ± 0.18 | 0.60 ± 0.22 | 0.52 ± 0.24 | 0.53 ± 0.165 | 0.57 ± 0.21 | 0.51 ± 0.14 |
| **Neutrophils % [1]** | 0.693 ± 0.11 | 0.769 ± 0.081 | 0.592 ± 0.083 | 0.659 ± 0.104 | 0.740 ± 0.096 | 0.600 ± 0.082 |
| **Lymphocyte % [1]** | 0.182 ± 0.99 | 0.113 ± 0.063 | 0.286 ± 0.072 | 0.220 ± 0.1 | 0.133 ± 0.081 | 0.278 ± 0.074 |
| **Monocyte % [1]** | 0.076 ± 0.021 | 0.067 ± 0.023 | 0.086 ± 0.018 | 0.078 ± 0.020 | 0.071 ± 0.023 | 0.084 ± 0.017 |
| **RBC [1E12/L]** | 4.37 ± 0.43 | 4.16 ± 0.43 | 4.60 ± 0.34 | 4.43 ± 0.425 | 4.20 ± 0.47 | 4.59 ± 0.35 |
| **Hb [g/L]** | 132 ± 13.5 | 124 ± 13 | 140 ± 11.5 | 133 ± 14 | 125 ± 14 | 140 ± 11 |
| **Hct [1]** | 0.393 ± 0.039 | 0.372 ± 0.037 | 0.418 ± 0.03 | 0.399 ± 0.04 | 0.375 ± 0.042 | 0.418 ± 0.031 |
| **MCV [fL]** | 90.2 ± 3.65 | 89.6 ± 3.65 | 91.0 ± 3.45 | 90.6 ± 3.7 | 89.7 ± 3.75 | 91.3 ± 3.5 |
| **MCH [pg/cell]** | 30.2 ± 1.25 | 29.9 ± 1.3 | 30.4 ± 1.2 | 30.3 ± 1.25 | 30.0 ± 1.4 | 30.5 ± 1.2 |
| **MCHC [g/L]** | 333 ± 7 | 333 ± 8 | 333 ± 6.5 | 333 ± 7 | 332 ± 7.5 | 333 ± 6 |
| **RDW [1]** | 0.138 ± 0.011 | 0.143 ± 0.012 | 0.133 ± 0.007 | 0.137 ± 0.01 | 0.143 ± 0.013 | 0.133 ± 0.007 |
| **Platelet count [1E9/L]** | 210 ± 54 | 227 ± 64 | 193 ± 42 | 203 ± 51 | 221 ± 59 | 191 ± 44 |
| **MPV [fL]** | 8.5 ± 0.75 | 8.5 ± 0.75 | 8.5 ± 0.7 | 8.5 ± 0.75 | 8.5 ± 0.8 | 8.5 ± 0.75 |
| **CRP [mg/L]** | 23 ± 54 | 90 ± 73.5 | 3 ± 3 | 6 ± 33.5 | 67 ± 82.5 | 3 ± 2.5 |
| **NLR [1]** | 3.80 ± 3.26 | 6.77 ± 4.74 | 2.07 ± 0.87 | 3.00 ± 2.39 | 5.62 ± 4.76 | 2.17 ± 0.90 |

additionally corrected using the Bonferroni correction [42]. As we can see from Table 2, the XGBoost model (XGB) has both a significantly higher classification accuracy (83.5 % ± 1 %) and a significantly lower Brier score (0.123 ± 0.007) than any other compared models (p < 0.01 in all cases, see Table S1), and was therefore used as the principal Virus vs. Bacteria model. Additional characteristics (ROC curves) of the final model are shown in Supplemental Fig. S2.

*3.5. CRP is of little diagnostic value in the range of 10–40 mg/L*

Based on Fig. 5(A and B) and Fig. 6 and for the purpose of our study, we define the CRP range of interest between 10 and 40 mg/L. In the training set, the CRP region between 10 and 40 mg/L includes 3636 cases (*14.6 %*). Of these 1764 ($p_B$=*48.5 %*) are labeled as 'Bacteria' and 1872 ($p_V$=*51.5 %*) as 'Virus'.

For all data, we tested the statistical significance of results using the one-sided Wilcoxon signed rank test and the paired *t*-test, both with the Bonferroni correction (Tables 4 and 5), the XGBoost model, has both a significantly (p < 0.01) higher classification accuracy (76 % ± 2.5 %) and a significantly lower (p < 0.01) Brier score (0.171 ± 0.016) than any other compared model, with the exception of the Random Forest (RF) model, where the difference in accuracy is not significant (p > 0.05). This result further justifies the choice of XGBoost as the principal Virus vs. Bacteria model.

In the CRP 10–40 mg/L region, the optimal CRP decision rule (see the Quantification and statistical analysis-Baseline approach section), diagnoses all patients with a CRP level lower than 24 mg/L as having a viral infection and all patients with higher CRP levels as having a bacterial infection. The CRP decision rule achieves an accuracy of 55.1 % ± 2 % (2003 out of 3636). Compared with two simplistic approaches (most prevalent and random diagnosis), this result is not impressive. Diagnosing every case completely at random yields the accuracy $p_B^2 + p_V^2 = 50.5$ %, while labeling every case as 'Virus' (the prevalent diagnosis in this region) yields an accuracy of 51.5 %. We can conclude that in the 10–40 mg/L range, the CRP decision rule is only slightly better than a random diagnosis (by 4.6 %) and the most prevalent diagnosis (by 3.6 %). On the other hand, the XGB model achieves an accuracy of 76 % ± 2.5 % (2764 out of 3636), 20.9 % ± 3.2 % better than the CRP decision rule (see Table 3 and Table S2).

The diagnostic significance of other blood parameters within this region is effectively elucidated through an examination of the importance of these parameters for the model across varying ranges of CRP, as depicted in Fig. 6.

*3.6. Model performance on the evaluation dataset*

For rigorous methodological integrity, the evaluation dataset, encompassing 6176 cases, was exclusively employed to assess the performance of the ultimate diagnostic model. The outcomes parallel those derived from 10-fold cross-validation: an accuracy of 82.2 % with 95 % binominal confidence intervals using the Agresti-Coull method [60] of ±1 %, a Brier score of 0.129, and an area under the ROC curve (AUC) of 0.905 (Table 4). The XGBoost model is again significantly superior to the CRP decision rule, both in terms of diagnostic accuracy and Brier score, especially in the CRP range between 10 and 40 mg/L (p < 0.01 for all comparisons, see Table S3). The performance of the optimal CRP decision rule (see the Quantification and statistical analysis-Baseline approach section), which diagnoses all patients with CRP levels lower than 24 mg/L as having a viral infection and all patients with higher CRP levels as having a bacterial infection, has an accuracy of 77.1 % with 95 % binominal confidence intervals using the Agresti-Coull method of ±1 %, a Brier score of 0.175, and an AUC of 0.818 (Table 4).

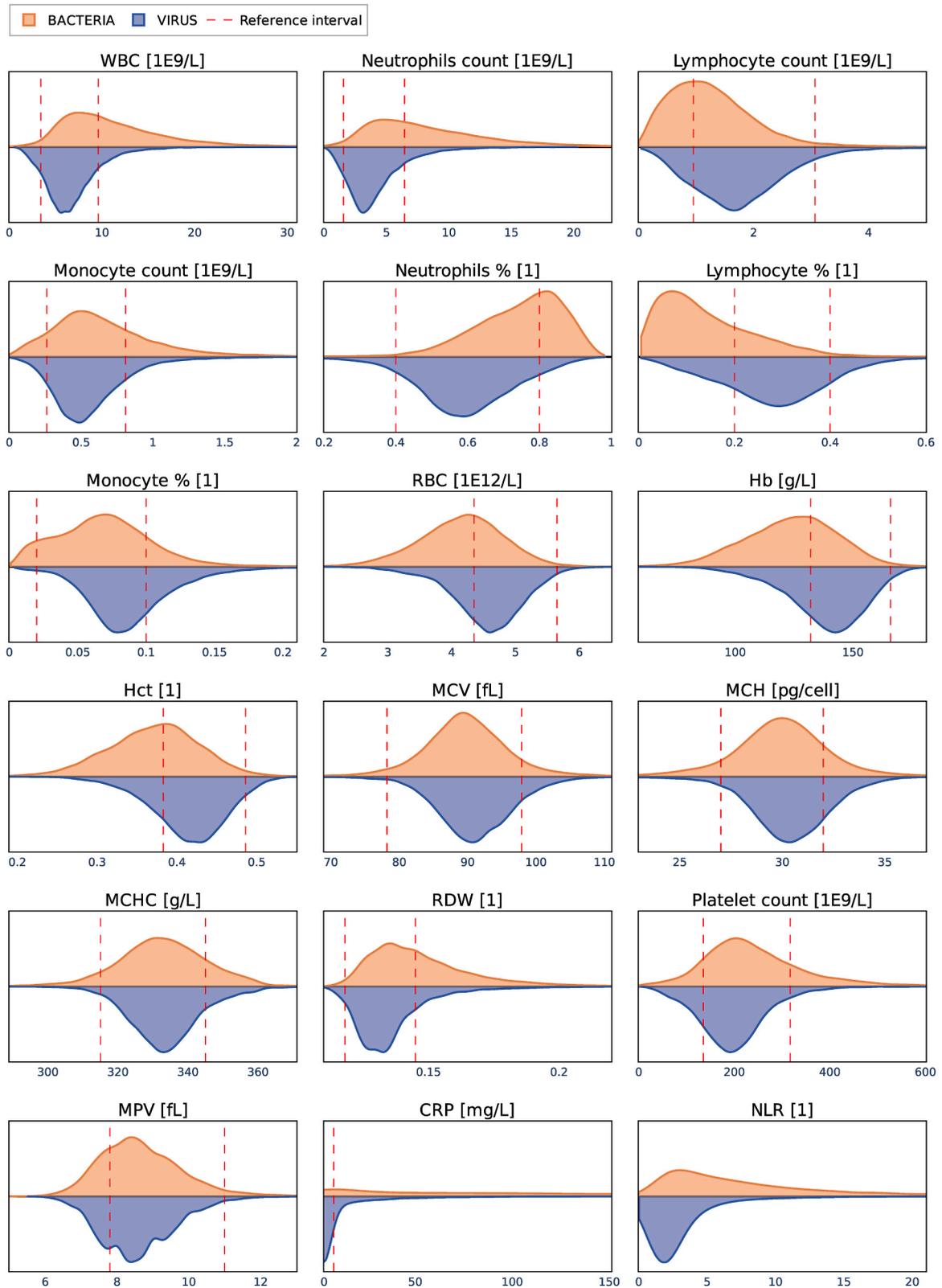

**Fig. 3.** Violin plots of blood parameters for visual comparisons of 'Bacteria' and 'Virus' populations. Most parameters exhibit considerable perceptive differences between the populations. For this visualization, all cases from the training set were included. The red vertical dashed lines represent reference intervals. (For interpretation of the references to color in this figure legend, the reader is referred to the Web version of this article.)

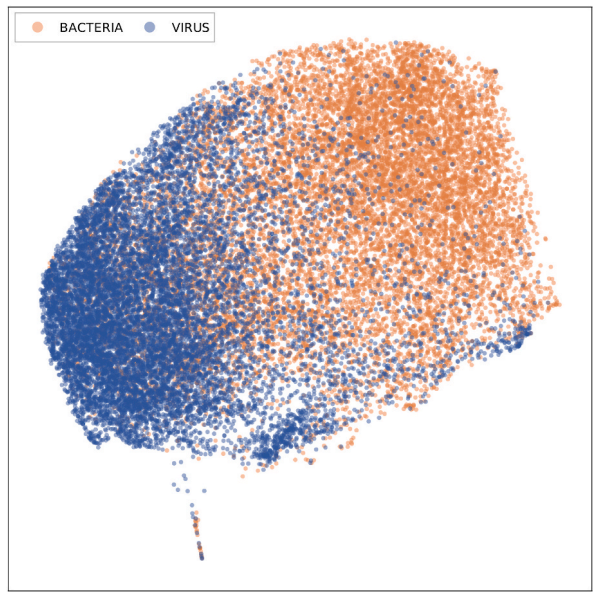

**Fig. 4.** Visualization of the parameter space with the UMAP method. Each dot represents a single blood test or, more specifically, an embedding of all blood parameters into a two-dimensional space, and its color represents the infection type. Blue dots represent blood tests with viral infections, and orange dots represent blood tests with bacterial infections. The visualization shows all the labeled cases from the training set. (For interpretation of the references to color in this figure legend, the reader is referred to the Web version of this article.)

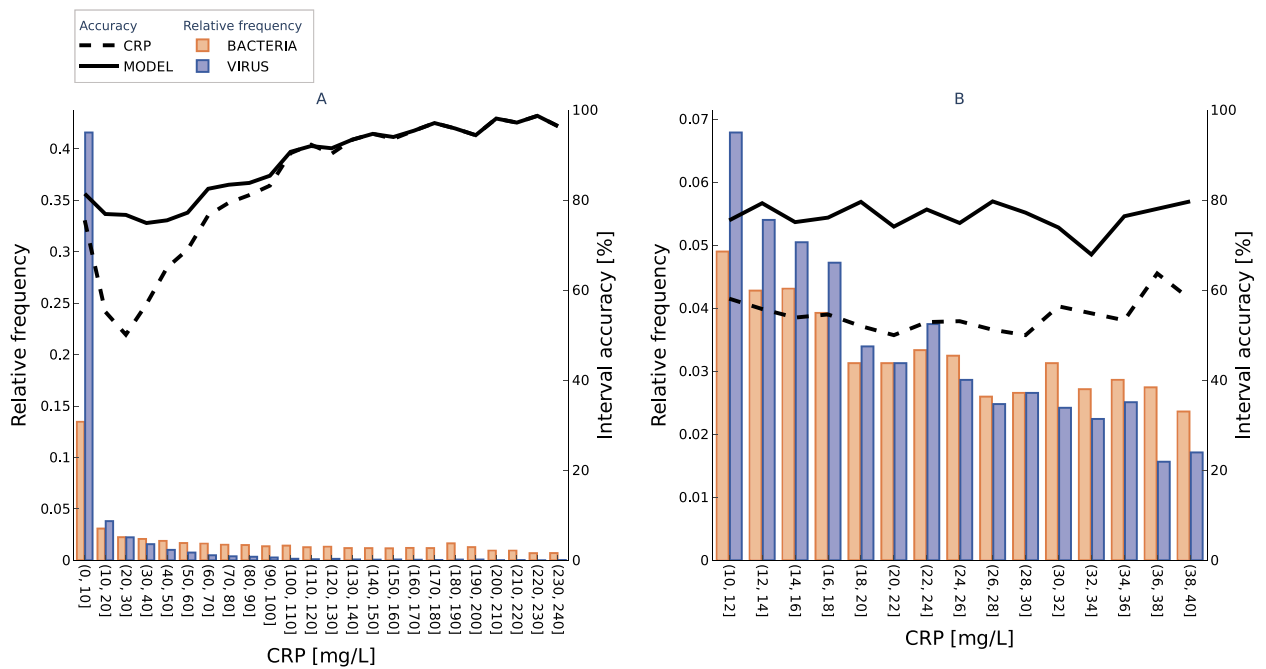

**Fig. 5.** Performance of the Virus vs. Bacteria model (model) and simple CRP decision rule (CRP) (A) on all cases from 10-fold cross-validation, (B) on all cases from 10-fold cross-validation that have CRP values within the region of interest between 10 and 40 mg/L.

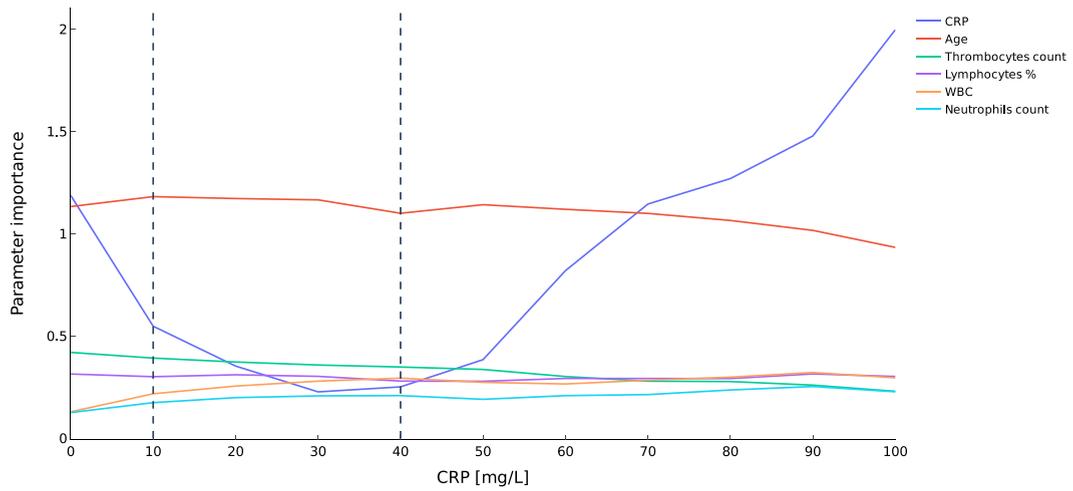

**Fig. 6.** Importance of the top six parameters as a function of different CRP concentration ranges. The vertical dashed lines denote the CRP range of 10–40 mg/L.

In the evaluation set, the CRP region between 10 and 40 mg/L includes 884 cases (*14.3 %*), with 428 ($p_B$=*48.4 %*) labeled as 'Bacteria' and 456 ($p_V$=*51.6 %*) labeled as 'Virus'. Here, the CRP decision rule is unable to distinguish between bacterial and viral infections (the ROC curve is close to the diagonal, and the AUC is only slightly above 0.5) (Fig. 7(A and B)). On the other hand, the Virus vs. Bacteria model exhibits a normal, useful ROC curve, similar to the result using all evaluation cases, and an AUC of 0.822, only 0.083 less than the result using all evaluation data (0.905). In this range, the Virus vs. Bacteria model achieved a diagnostic accuracy of 73.2 % ± 2.9 %, which is 17.9 % ± 4.1 % more accurate than the CRP decision rule (55.3 % ± 3.3 %) (Table 4).

### 3.7. Parameter importance to the model

Shapley values were calculated for each blood parameter to understand its contribution to the XGB model's decision-making process. CRP contributes the most, followed by platelet count, lymphocyte%, WBC count, neutrophils count, etc. (Fig. 8). Together with the parameter data analysis (Table 1 and Fig. 3), we can find the most medically relevant parameters for distinguishing between bacterial and viral infections to be CRP, WBC, neutrophil count, lymphocyte count, platelet count, and age.

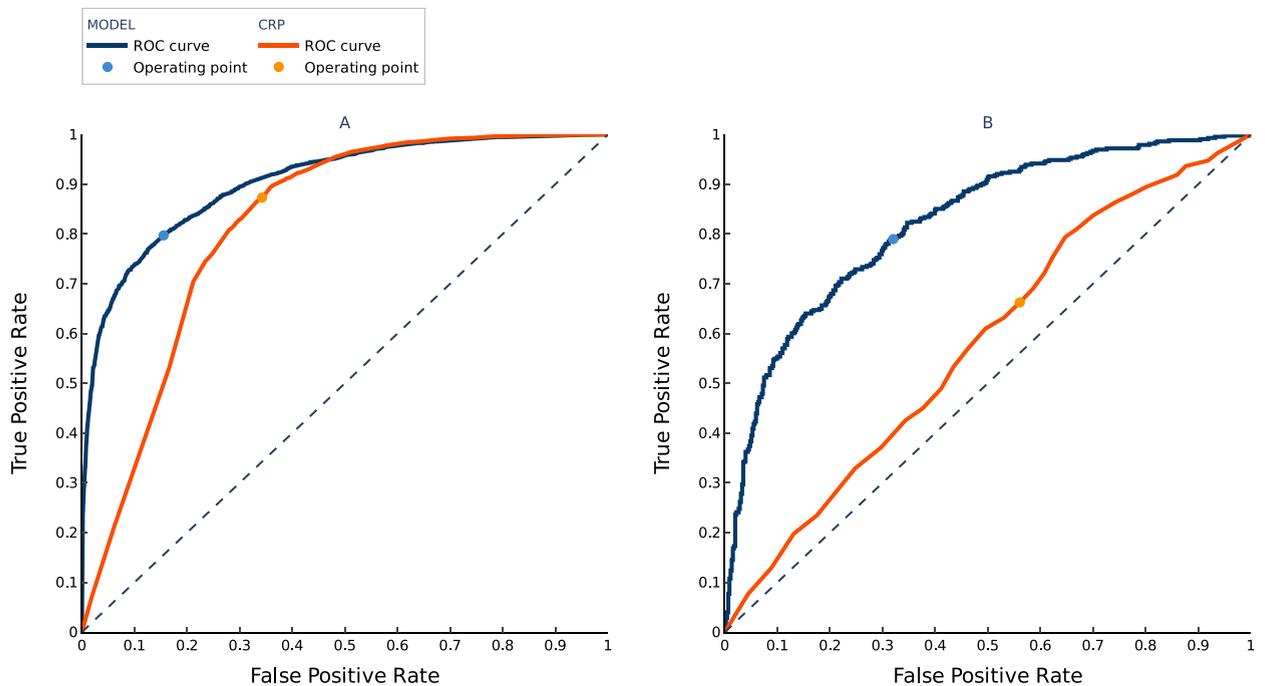

**Fig. 7.** ROC curves of the Virus vs. Bacteria model (model) and simple CRP decision rule (CRP) (A) on all evaluation cases, (B) on all evaluation cases that have CRP values within the region of interest between 10 and 40 mg/L. For the model, a default operating point (0.5) is visualized; for the CRP decision rule, an operating point of 24 mg/L is displayed.

In Table 1 we can see considerable median age differences between populations of 'Virus' and 'Bacteria' cases. To address this possible source of bias, we built another model using the same dataset but excluding age and compared the results. Overall, the model without age is only slightly worse (by 2 %) than the final Virus vs. Bacteria model and better than the CRP decision rule (by 4 %). Fig. S1 shows a comparison of the model performance stratified by age. We can see that while knowing the patient age helps the model, the difference in accuracy by strata is mostly in the range of 2 %.

## 4. Discussion

This study demonstrates the potential of machine learning models, particularly the Virus vs. Bacteria model, based on the most frequently measured blood parameters and CRP, in differentiating between bacterial and viral infections. Differentiating the main types of infections is an essential development in the ongoing efforts to combat antibiotic resistance, as more accurate diagnostic methods are crucial for improving patient care and reducing unnecessary antibiotic usage. Luz et al. [61] analyzed 158,616 journal articles related to antimicrobial resistance published between 1999 and 2018. They show that the number of antimicrobial resistance-related publications has grown 450 % over the two decades.

Our study builds on our previous studies that have leveraged machine learning using blood test results for differential diagnosis in hematology [21], brain tumors [23] and particularly in COVID-19 [22]. Several other studies have explored the use of different, more specific blood biomarkers, to distinguish between bacterial and viral infections and also the use of machine learning in this process. Oved and colleagues [62,63] developed an assay combining three host proteins: tumor necrosis factor-related apoptosis-inducing ligand (TRAIL), interferon gamma induced protein-10 (IP-10) and CRP (AUC of 0.94). The assay was significantly more accurate in febrile children than CRP, procalcitonin, and routine laboratory parameters [6]. De Jager [54] showed that the neutrophil-lymphocyte count ratio (NLR) in the emergency department predicts the severity and outcome of community-acquired pneumonia with higher prognostic accuracy than traditional infection markers, and an additional utility in elderly adults was demonstrated by Cataudella [55]. NLR has also been identified as a diagnostic and prognostic tool in viral infections, including COVID-19 [57,58]. Shen et al. [64] presented an ontology-driven clinical decision support system IDDAP for infectious disease diagnosis and antibiotic prescription. It includes a rule-based inference engine that applies ontology-based rules to patient data and a user interface that presents the results of the inference to the clinician. IDDAP was evaluated using a dataset of 84 patients, yielding an AUC of 0.899, thus showing the potential to improve patient outcomes and reduce healthcare costs. Ramgopal et al. derived an ML model to predict serious bacterial infections in young febrile infants [65]. Lien et al. [66] have recently shown that an ML model using only complete blood count with differential leukocyte count data achieved similar performance as procalcitonin data in bacteremia prediction and Li et al. [67], showed in 293 patients with suspected lower respiratory tract infections and/or sepsis that the most predictive variable was CRP. New experimental methods for viral/bacterial classification also build on ML methods; a nice example is the use of ML on infrared spectroscopy to diagnose inaccessible infections [68,69]. These studies highlight the potential of machine learning and biomarker-based approaches in improving diagnostic accuracy.

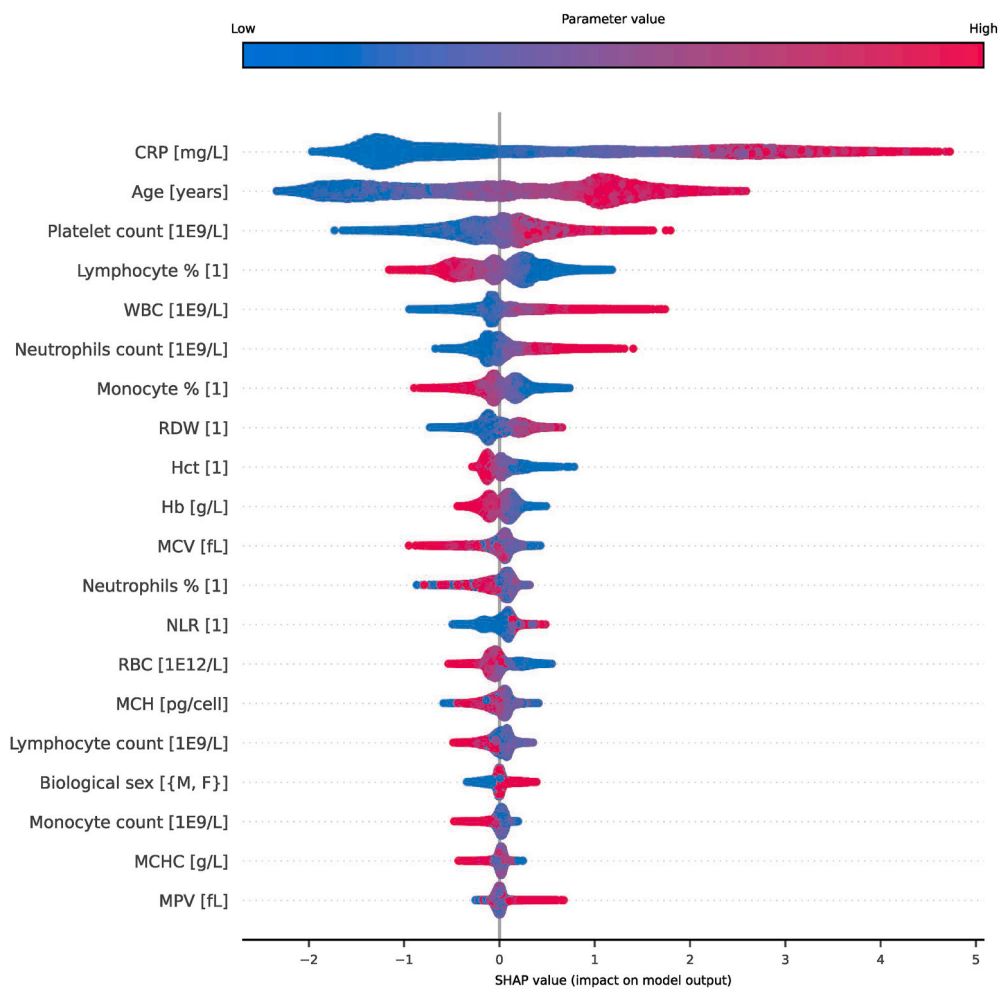

**Fig. 8.** Distribution of Shapley values across the entire dataset, which can be used to identify the features that have the greatest impact on the model's predictions. Each dot represents a case in the dataset, and its position along the x-axis corresponds to the Shapley value for a particular blood parameter. The color of the dot indicates the value of the feature for that instance, with blue indicating a low value and red indicating a high value. The position of the dot indicates its impact on the prediction: dots to the left of the zero vertical line indicate a negative impact, while dots to the right indicate a positive impact. The larger the absolute value of the Shapley value, the larger the impact of the corresponding feature on the prediction. (For interpretation of the references to color in this figure legend, the reader is referred to the Web version of this article.)

Several studies have proposed different CRP cutoff values for prescribing antibiotics. In a Danish study [48], 59 % of patients with acute respiratory tract infection had a CRP test performed. At least 25 % of the patients were prescribed an antibiotic when the CRP level was >20 mg/L, 50 % when the CRP was >40 mg/L, and 75 % when the CRP was >50 mg/L. In another study [49] among 372 patients with acute cough tested with a POCT for CRP, the CRP value was the strongest independent predictor of antibiotic prescription, with an odds ratio of CRP $\geq$ 50 mg/L of 98.1. There are also recommendations that for CRP<20 mg/L, the prescription of antibiotics should be delayed [47,48]. On the other hand, normal CRP levels are usually considered to be below a fixed threshold, ranging from 3 to 10 mg/L. Consequently, relying solely on CRP levels for clinical decision-making may lead to misdiagnoses and inappropriate antibiotic prescriptions.

One of the main findings of our study is the superior performance of the Virus vs. Bacteria model compared to the CRP decision rule, which is likely a close approximation to the way a physician decides whether the infection is of a viral or bacterial origin. This indicates that routinely measured blood parameters have important diagnostic value in distinguishing viral and bacterial infections. Moreover, the Virus vs. Bacteria model showed a remarkable improvement in accuracy within the CRP range of 10–40 mg/L, a region where the CRP decision rule has limited diagnostic value. Our analysis shows that using a simple CRP based decision rule outside of the 10–40 mg/L CRP range is almost on par with our model. However, even outside of this CRP range, there are some bacterial infections in our dataset with a relatively low CRP, like spirochetal infections (Lyme disease, syphilis, leptospirosis) with a median CRP of 5 mg/L, and viral infections with a relatively high CRP, like viral pneumonia and influenza with a median CRP around 65 mg/L) and in COVID-19 with a median CRP of 42 mg/L. Spirochetes can evade the host's immune system, leading to lower CRP levels during infection because the host's immune response relies more on cellular immunity rather than inflammation [70,71], which can be detected with other blood parameters. On the other hand, viral infections that affect the respiratory system cause inflammation in the airways and lung tissue, which leads to the production of CRP [72]. This can also be an indication of severe infection or complications, such as secondary bacterial infections or a more pronounced inflammatory response [22,73]. This further highlights the benefits of a multifactorial approach to diagnosing infections.

The Shapley value analysis (Fig. 8) offered valuable insights into the relative contributions of individual blood parameters to the Virus vs. Bacteria model's decision-making process. CRP, WBC, neutrophils count, lymphocyte count, and platelet count emerged as the most medically relevant blood parameters in distinguishing between bacterial and viral infections. This knowledge can help clinicians prioritize these critical parameters when examining patient data. Elevated WBC counts often indicate bacterial infections, as the immune system combats the infection by increasing the production of these cells [74]. In contrast, viral infections usually lead to lower WBC counts or a modest increase due to the direct impact on bone marrow, immune-mediated destruction or redistribution of white blood cells, and the body's stress response. Neutrophil count differences between bacterial and viral infections can be explained by their unique roles in immune response. Bacterial infections typically involve increased neutrophil counts due to rapid recruitment for bacterial elimination via phagocytosis [75]. In contrast, viral infections often decrease neutrophil count, as the immune system favors deploying lymphocytes, particularly T cells, to combat viruses, leading to a relative increase in differential analysis. Distinct differences in platelet count have been observed between bacterial and viral infections. Bacterial infections typically present higher platelet counts due to direct interaction and activation of platelets with bacteria [76]. In contrast, viral infections can cause thrombocytopenia, a reduced platelet count, due to impaired platelet production or increased destruction induced by the virus [77].

Besides blood parameters, age is an important parameter as it is a critical determinant due to the alterations in hematopoiesis associated with the aging process. Age is also correlated with a rise in bacterial infections and a decline in viral infections. (Fig. 8, age). This is likely due to immunosenescence and a higher prevalence of comorbidities [41,78]. Lower susceptibility to viral infections in older individuals might be attributed to prior exposure, which enhances immune memory and protection [79]. However, the severity and clinical outcomes of viral infections can be more severe in older individuals due to age-related decline of immune function [80].

This study demonstrates the potential of machine learning, specifically the Virus vs. Bacteria model, in accurately differentiating between bacterial and viral infections based on routinely measured blood parameters and CRP. The superior performance of the Virus vs. Bacteria model over the CRP decision rule highlights the importance of considering multiple blood parameters in diagnostic decision-making. The insights gained from this study may contribute to the development of more sophisticated diagnostic tools that leverage machine learning and relevant biomarkers to facilitate better clinical decision-making in the management of infections.

Future research should focus on validating these results in larger and more diverse patient cohorts as well as exploring alternative machine learning models to further improve diagnostic accuracy. This might also be achieved by monitoring temporal changes in blood parameters, potentially enhancing accuracy with a reduced number of measured parameters. Additionally, there's potential in simplifying the model, especially for the CRP 10–40 mg/L range. This could be accomplished by incorporating additional parameter ratios (beyond NLR) and reconfiguring the model into more straightforward decision rules, thereby facilitating its application even in settings without advanced computational resources.

*4.1. Limitations of the study*

The current study only considered confirmed ICD-encoded diagnoses, and some patients could not be definitively classified as having bacterial or viral infections. These cases were used for semi-supervised training, and their inclusion may have introduced some bias in the model's performance. Our objective was to distinguish between bacterial and viral infections, prompting us to exclude cases exhibiting both types. The model, which has been trained exclusively on data from patients diagnosed with bacterial or viral infections, lacks representation from healthy individuals. This limitation confines the model's applicability to populations already exhibiting infection symptoms and consequently renders it less suitable for broader screening purposes where a mix of healthy and infected individuals is expected. Additionally, we did not include specific biomarkers like procalcitonin, TRAIL, IP-10, and others, despite their promising results in similar studies. Our analysis aimed to demonstrate the utility of standard blood tests, which may still be overly comprehensive for certain clinical settings. The data for this study was sourced from a single medical center, suggesting the need for future research to explore the findings' applicability in varied clinical environments.

**Funding**

This research and development of the Virus vs. Bacteria model were supported by Smart Blood Analytics Swiss SA.

**Data availability statement**

Data: The data contains sensitive personal information and hence cannot be made publicly available. Any additional information is available from the lead contact upon request.

The National Ethics Committee of Slovenia approved the study: No. 0120–718/2015/7, No. 0120–170/2020/6, No. 0120-058/2016-2 KME 7/01/16, No. 0120–341/2016-2 KME 33/07/16, No. 0120–718/2015-2 KME 103/11/15 and No. 0120–718/2015/5.

Software: The Virus vs. Bacteria model (VB_20230104_75333_19) is available at https://app.smartbloodanalytics.com under the Terms & Conditions and can be utilized upon registration and reasonable request for its use.

Code: The pseudo codes of the semi-supervised bootstrap labeling and SBAS machine learning pipeline are included within supplemental materials (Algorithms 1 and 2).

**CRediT authorship contribution statement**

**Gregor Gunčar:** Writing – review & editing, Writing – original draft, Supervision, Project administration, Methodology, Investigation, Formal analysis, Conceptualization. **Matjaž Kukar:** Writing – review & editing, Writing – original draft, Visualization, Validation, Supervision, Software, Methodology, Investigation, Formal analysis, Data curation, Conceptualization. **Tim Smole:** Writing – review & editing, Writing – original draft, Visualization, Validation, Software, Methodology, Investigation, Formal analysis, Data curation. **Sašo Moškon:** Writing – review & editing, Writing – original draft, Visualization, Validation, Software, Methodology, Investigation, Formal analysis, Data curation. **Tomaž Vovko:** Investigation, Data curation. **Simon Podnar:** Writing – review & editing, Investigation, Data curation. **Peter Černelč:** Writing – review & editing, Investigation, Data curation. **Miran Brvar:** Writing – review & editing, Investigation, Data curation. **Mateja Notar:** Writing – review & editing, Project administration, Data curation. **Manca Köster:** Writing – review & editing, Data curation. **Marjeta Tušek Jelenc:** Writing – review & editing. **Žiga Osterc:** Software, Project administration, Data curation. **Marko Notar:** Writing – review & editing, Writing – original draft, Supervision, Project administration, Investigation, Funding acquisition, Data curation, Conceptualization.

**Declaration of competing interest**

Marko Notar is the CEO of Smart Blood Analytics SA. Mateja Notar, Sašo Moškon, Tim Smole, Žiga Osterc, Marjeta Tušek Jelenc and Manca Köster hold positions at Smart Blood Analytics Swiss SA. Matjaž Kukar, Peter Černelč, and Gregor Gunčar serve as advisors to Smart Blood Analytics Swiss SA. The remaining authors declare no competing interests.

**Appendix A. Supplementary data**

Supplementary data to this article can be found online at https://doi.org/10.1016/j.heliyon.2024.e29372.